\documentclass[letterpaper, 10 pt, conference]{ieeeconf}  

\IEEEoverridecommandlockouts                              

\usepackage{graphics} 
\usepackage{epsfig} 
\usepackage{mathptmx} 
\usepackage{times} 
\usepackage{amsmath} 
\usepackage{cleveref}
\usepackage{amssymb}
\usepackage{multirow}
\usepackage{booktabs}
\usepackage{cite}
\usepackage{authblk}

\title{\LARGE \bf
Uncertainty-Aware Hierarchical Re-Localization \\in OpenStreetMap via Semantic Alignment
}

\author{Yuchen Zou, Xiao Hu, Lihuang Fang, Yuqing Tang*
\thanks{This work was completed during Y. Zou's internship at the International Digital Economy Academy.}
\thanks{Y. Zou is with the School of Automation Science and Engineering, Xi'an Jiaotong University, Xi'an, Shaanxi 710049, China.}
\thanks{X. Hu and Y. Tang are with the International Digital Economy Academy, Guangdong, Shenzhen 510085, China.}
\thanks{L. Fang is with the Department of Electronic and Electrical Engineering, Southern University of Science and Technology, Shenzhen 518055, China.}
\thanks{* Corresponding authors (e-mail: tangyuqing@idea.edu.cn).}
}

\begin{document}

\maketitle
\thispagestyle{empty}
\pagestyle{empty}

\begin{abstract}

Monocular re-localization enables robots to estimate camera poses from visual observations. However, many existing methods rely on dense maps or large reference image databases, which face scalability limitations and privacy risks. OpenStreetMap (OSM), as a lightweight privacy-preserving map, offers semantic and geometric information with global scalability. Nonetheless, OSM localization remains challenging due to cross-modal discrepancies between natural images and OSM, as well as the high cost of global map-based localization. In this paper, we propose an uncertainty-aware hierarchical search framework with semantic alignment for localization in OSM. First, object-centric DINO-ViT tokens are exploited to reduce the semantic gap between ground-view observations and OSM vectors. Second, global dense matching is decomposed into coarse FFT correlation and uncertainty-controlled local refinement. Extensive experiments demonstrate that our method significantly improves localization accuracy and speed. When trained on a single dataset, the 3$^\circ$ orientation recall of our method even outperforms the 5$^\circ$ recall of state-of-the-art methods.

\end{abstract}

\section{Introduction}

Monocular visual re-localization estimates camera pose from a single image, serving as a core technology for autonomous driving \cite{c3}, navigation \cite{c2}, and virtual reality \cite{c4,c5,c6}. Compared to LiDAR and multi-camera or stereo systems, monocular systems offer advantages in hardware cost and deployment flexibility, making them scalable for commercial applications.

Traditional monocular localization often relies on dense maps \cite{c7} or extensive reference image databases \cite{c8, c9}. These representations face scalability issues due to construction and maintenance costs \cite{c10}, alongside privacy leakage risks during data acquisition \cite{c11}. Consequently, OpenStreetMap (OSM) has emerged as a compelling alternative. As an open-source vector map, OSM continuously updates building footprints and road topologies \cite{c12}, providing rich, privacy-friendly semantic and geometric priors for localization \cite{c13}.

Despite its advantages, accurate localization on OSM remains extremely challenging. Monocular images capture perspective surface textures, illumination, and viewpoint variations, whereas OSM provides structured, top-down geometric abstractions. This severe modality gap makes direct feature matching highly ineffective. Furthermore, performing exhaustive fine-grained search across large-scale OSM regions incurs prohibitive computational costs, failing to meet the real-time requirements of robotic systems \cite{c14}.

\begin{figure}[t]
\centering
\includegraphics[width=0.9\columnwidth]{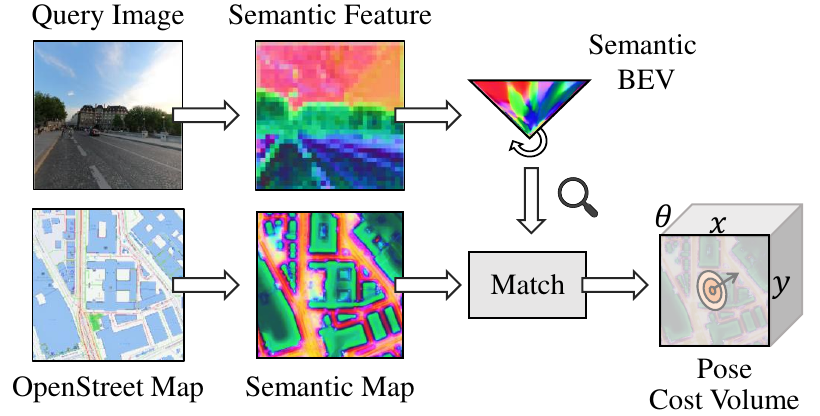}
\caption{Localization in the OSM-derived semantic map by transforming the query image into semantic features.}
\label{fig1}
\end{figure}

To address these hurdles, we re-formulate OSM-based localization as a cross-modal semantic alignment problem. Leveraging the object-centric semantic tokens of DINO-ViT \cite{c15}, we decouple complex ground-view features to align with the vector semantics and geometric elements of OSM (as shown in Fig. \ref{fig1}). Moreover, we depart from traditional global dense matching paradigms \cite{c14, c16} by decomposing search into coarse FFT correlation and uncertainty-aware local refinement. This hierarchical approach progressively refines the location and heading direction, reducing computational overhead while ensuring high registration accuracy.

The main contributions are summarized as follows:
\begin{itemize}
    \item We exploit object-centric DINO-ViT tokens to bridge the cross-modal gap between ground-view observations and OSM vector semantics.
    \item We propose an uncertainty-aware coarse-to-fine localization mechanism that converts global dense matching into coarse FFT correlation and adaptive local refinement.
    \item Experiments on street-view and autonomous driving datasets demonstrate that this mechanism triples localization speed. Remarkably, when trained on a single dataset, our 3$^\circ$ orientation recall outperforms the 5$^\circ$ recall of state-of-the-art methods.
\end{itemize}

\begin{figure*}
\centering
\includegraphics[width=1.6\columnwidth]{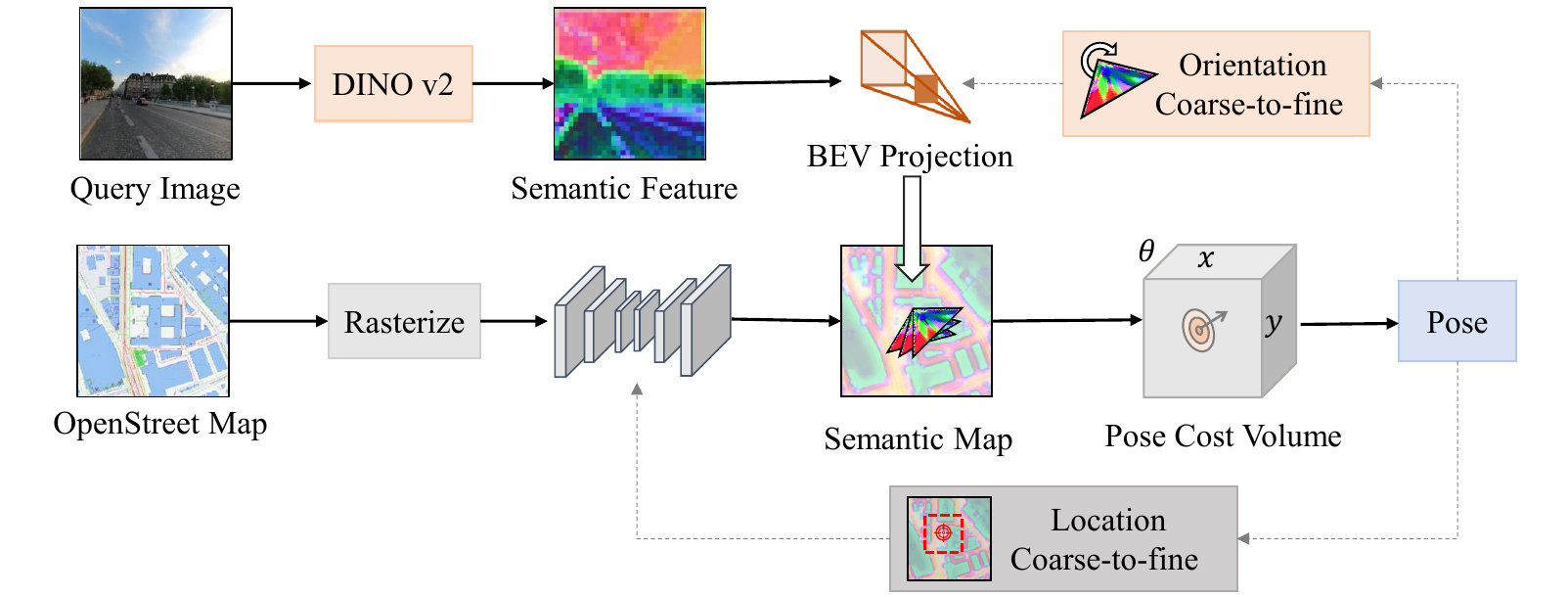}
\caption{Overview of the proposed coarse-to-fine OSM localization framework.}
\label{fig3}
\end{figure*}

\section{Related Work}

\subsection{Monocular Re-Localization in OSM}
Monocular re-localization has evolved from image retrieval \cite{c18, c19, c20} and map-based visual localization \cite{c17, c21} toward leveraging OpenStreetMap (OSM) \cite{c22} to circumvent high maintenance costs and privacy issues. As a lightweight vector database, OSM provides structured geometric priors like road networks and building footprints \cite{c12, c23, c24, c25}. Although frameworks like OrienterNet \cite{c14} use OSM to build semantic neural maps, they struggle with the modality gap between ground-level perspective views and overhead vector abstractions. Furthermore, the computational burden of dense matching across extensive OSM regions limits real-time deployment.

\subsection{Hierarchical and Neural Map-based Localization}
To balance efficiency and precision, hierarchical localization typically adopts a progressive coarse-to-fine refinement paradigm. This approach utilizes global retrieval to provide an initial location hypothesis \cite{c27}, which is subsequently verified and refined through local feature matching \cite{c28, c29}. In the domain of neural maps, MapLocNet \cite{c30} attempts to regress camera poses directly from heterogeneous map data. Nevertheless, direct regression frequently lacks the necessary robustness when addressing the high-frequency geometric variations found in single-view imagery. In contrast, our framework employs a matching-based hierarchical strategy. This method preserves the interpretability of geometric alignment while benefiting from the optimization capabilities of differentiable neural cost fields.

\subsection{Semantic Representation via Vision Transformers}
Vision Transformers (ViT) \cite{c31}, particularly self-supervised models like DINO \cite{c15, c43}, extract robust, object-centric representations invariant to lighting or texture fluctuations \cite{c37, c38}. Unlike convolutional features that prioritize local textures, the tokens generated by DINO are aware of global context and naturally decouple essential urban elements such as building facades and road boundaries. These semantic primitives correspond directly to the area and line features defined within OSM. By adopting DINO as a semantic backbone, we establish a robust intermodal bridge that aligns visual observations with geometric constraints from the map, ensuring strong generalization in complex, unconstrained environments.


\section{Method}

Given a query image $I$ and a candidate local OSM map crop, we aim to estimate the 3-DoF camera pose $\xi = (x, y, \theta)$ within the map crop. Our framework aligns both the monocular observation and the local OSM map crop within a unified semantic space to perform robust registration (Fig. \ref{fig3}).

\subsection{Semantic BEV Projection}
We employ DINO to extract high-level semantic features $F_I \in \mathbb{R}^{U \times V \times N}$ from the query image $I$. To preserve pretrained semantics while adapting high-level features to localization, we freeze the first 8 layers of the DINO encoder and fine-tune the final 4 layers.

Subsequently, $F_I$ is projected into the Bird's-Eye View (BEV) space using a monocular scale-distribution method \cite{c14}. Specifically, $F_I$ is first mapped to polar coordinates $F_{polar} \in \mathbb{R}^{U \times D \times N}$ via depth-based ray interpolation:
\begin{equation}
F_{polar}(u,d,n) = \sum_{v} \alpha_{(d,v)} \cdot F_I(u,v,n),
\end{equation}
where $\alpha_{(d,v)}$ denotes the interpolation probability between image column $v$ and depth plane $d$. The polar feature is then horizontally resampled onto a Cartesian grid to yield the semantic BEV representation $F_{bev} \in \mathbb{R}^{L \times D \times N}$:
\begin{equation}
F_{bev}(l,d,n) = \sum_{u} \alpha_{(u,l)} \cdot F_{polar}(u,d,n),
\end{equation}
where $\alpha_{(u,l)}$ is the transformation ratio determined by the camera intrinsics.

\subsection{Semantic Map Encoding}
OSM provides structured vector elements classified into areas (buildings), lines (roads), and points (POIs). To enable dense feature matching, we crop a local OSM map and project these discrete categorical vectors into a continuous feature space via a learnable embedding layer. Following \cite{c30}, a U-Net architecture is employed to generate a rasterized neural semantic map $F_{map} \in \mathbb{R}^{W \times H \times N}$. This encoding ensures the channel dimension aligns with $F_{bev}$ while preserving topological priors.

\begin{figure*}[t]
\centering
\includegraphics[width=1.9\columnwidth]{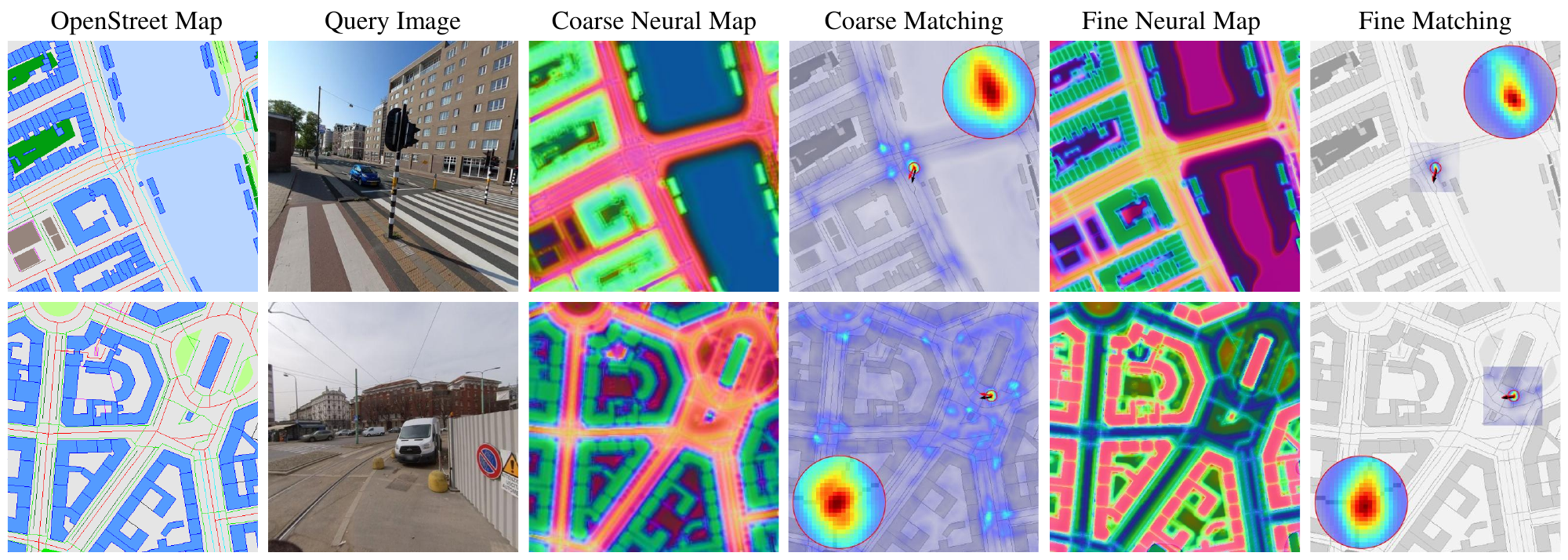}
\caption{Visualization of coarse-to-fine localization, including OSM neural-map refinement and pose-likelihood refinement. The red and black arrows denote the predicted and ground-truth poses.}
\label{fig4}
\end{figure*}

\subsection{Coarse-to-Fine Pose Estimation}
Exhaustive global dense matching incurs prohibitive computational costs. To balance efficiency and accuracy, we implement a hierarchical correlation strategy.

\textbf{Coarse Matching:} We generate a discretized set of rotationally equivariant BEV features using a coarse angular step $\Delta \theta^c$:
\begin{equation}
\tilde{F}_{bev}^c(\theta) = \mathbf{R}(\theta) \cdot F_{bev}^c, \quad \theta \in \{0, \Delta \theta^c, \dots, 2\pi-\Delta \theta^c\},
\end{equation}
where $\mathbf{R}(\theta) \in SO(2)$ is the rotation operator. Concurrently, low-resolution map features $F_{map}^c$ are extracted from the U-Net bottleneck. We compute a 3-DoF correlation cost field $M(p, \theta)$ via the Fourier domain:
\begin{equation}
M(\cdot, \theta) = \mathcal{F}^{-1} \left[ \mathcal{F}(F_{map}^c) \odot \mathcal{F}(\tilde{F}_{bev}^c(\theta))^* \right],
\end{equation}
where $\mathcal{F}$ and $\mathcal{F}^{-1}$ denote 2D Fast Fourier Transform and its inverse, respectively, and $*$ is the complex conjugate. The optimal coarse pose hypothesis $(p^c, \theta^c)$ is obtained via Maximum Likelihood Estimation evaluated at point $p$:
\begin{equation}
(p^c, \theta^c) = \arg\max_{(p \in P, \theta \in \Theta)} M(p, \theta).
\end{equation}

\textbf{Fine Refinement:} To adaptively constrain the high-resolution search space without compromising accuracy, we introduce a pose uncertainty measure $U$ evaluated over the local neighborhood $\mathcal{N}(p^c, \theta^c)$:
\begin{equation}
U = 1 - \frac{\exp(M(p^c, \theta^c))}{\sum_{(p, \theta) \in \mathcal{N}} \exp(M(p, \theta))}.
\end{equation}
This inverse-confidence metric dynamically contracts the fine-grained search space for orientation $\Theta^f$ and position $P^f$:
\begin{equation}
\Theta^f = [\theta^c - k_\theta U \Delta \theta^f, \theta^c + k_\theta U \Delta \theta^f],
\end{equation}
\begin{equation}
P^f = [p^c - k_p U, p^c + k_p U],
\end{equation}
where $k_\theta$ and $k_p$ are scaling factors, and $\Delta \theta^f$ is the base angular radius of the fine search window. As shown in Fig. \ref{fig4}, uncertainty-aware refinement removes redundant spatial queries while maintaining registration accuracy.

\subsection{Training Objective}
To enable end-to-end optimization of the semantic feature extractor and the neural map encoder, we formulate pose estimation as a probability density estimation problem. Given the ground truth pose $\xi_{gt} = (p_{gt}, \theta_{gt})$, we supervise both the coarse and fine matching stages using a Negative Log-Likelihood (NLL) loss. The total training objective is defined as:
\begin{equation}
\mathcal{L} = \mathcal{L}_{coarse}(\xi_{gt}, M^c) + \lambda \mathcal{L}_{fine}(\xi_{gt}, M^f),
\end{equation}
where $\lambda$ is a balancing hyperparameter. For each stage, the loss minimizes the negative log-probability of the ground truth pose, derived by applying a softmax function over the position-orientation correlation volume. Jointly optimizing this objective maximizes the correlation response at true geometric intersections and suppresses false-positive alignments in repetitive urban environments.


\begin{table*}
\centering
\caption{Localization performance on KITTI using satellite and OSM maps, with models trained on KITTI or MGL+KITTI.}
\label{tab2}
{
\begin{tabular}{llcccccccccc}
\toprule
\multirow{2}{*}{Map} & \multirow{2}{*}{Method} & \multirow{2}{*}{Train dataset} & \multicolumn{3}{c}{Lateral R@Xm} & \multicolumn{3}{c}{Longitudinal R@Xm} & \multicolumn{3}{c}{Orientation R@X$^\circ$} \\ \cmidrule(lr{0pt}){4-6} \cmidrule(lr{0pt}){7-9} \cmidrule(lr{0pt}){10-12}   &   &   & 1m & 3m & 5m & 1m & 3m & 5m & 1$^\circ$ & 3$^\circ$ & 5$^\circ$ \\ \toprule
\multirow{3}{*}{Satellite} & DSM \cite{c46}  & KITTI & 10.77 & 31.37 & 48.24 & 3.87 & 11.73 & 19.50 & 3.53 & 14.09 & 23.95 \\
                            & VIGOR \cite{c47} & KITTI & 17.38 & 48.20 & 70.79 & 4.07 & 12.52 & 20.14 & 5.46 & 15.79 & 27.15 \\
                            & Refinement \cite{c42} & KITTI & 27.82 & 59.79 & 72.89 & 5.75 & 16.32 & 26.48 & 18.42 & 49.72 & 71.00 \\ \toprule
\multirow{4}{*}{OSM}  & OrienterNet \cite{c14} & KITTI & 51.26& 84.77& 91.81& 22.39& 46.79& 57.81& 20.41& 52.24& 73.53\\
                               & Ours & KITTI & \textbf{68.05} & \textbf{92.84} & \textbf{95.81} & \textbf{27.45} & \textbf{57.89} & \textbf{68.34} & \textbf{37.28} & \textbf{80.16} & \textbf{92.12} \\ \cmidrule{2-12}
                            & OrienterNet \cite{c14} & MGL+KITTI & 65.91 & 92.76 & 96.54 & 33.07 & 65.18 & 75.15 & 35.72 & 77.49 & 91.51 \\
                            & Ours & MGL+KITTI &\textbf{72.28} & \textbf{94.42} & \textbf{97.49} & \textbf{37.87} & \textbf{71.40} & \textbf{79.61} & \textbf{42.83} & \textbf{87.26} & \textbf{96.37 }\\ \bottomrule
\end{tabular}
}
\end{table*}

\begin{table}
\centering
\setlength{\tabcolsep}{5pt}
\caption{Localization performance on MGL. 'C' and 'F' denote coarse and fine localization.}
\label{tab1}
{
\begin{tabular}{lccccccc}
\toprule
\multirow{2}{*}{Method} & \multicolumn{3}{c}{Position R@Xm} & \multicolumn{3}{c}{Orientation R@X$^\circ$} & \multirow{2}{*}{FPS} \\ \cmidrule(lr{0pt}){2-4} \cmidrule(lr{0pt}){5-7}
             & 1m    & 3m    & 5m    & 1$^\circ$    & 3$^\circ$    & 5$^\circ$    &       \\ \midrule
Retrieval       & 2.02  & 15.21 & 24.21 & 4.50  & 18.61 & 32.48 & -  \\
Refinement & 8.09  & 26.02 & 35.31 & 14.92 & 36.87 & 45.19 & -  \\
OrienterNet    & 15.78 & 47.75 & 58.98 & 22.14 & 52.56 & 66.32 & 7.51  \\
Ours-C  & 14.70  & 49.58 & 64.43 & 23.26 & 59.75 & 78.99 & \textbf{29.54} \\
Ours-F    & \textbf{17.36} & \textbf{53.67} & \textbf{65.81} & \textbf{30.66} & \textbf{68.71} & \textbf{80.53} & 25.80 \\
\bottomrule
\end{tabular}
}
\end{table}

\section{Experiments}

\subsection{Datasets and Implementation Details}
We evaluate our framework on two benchmarks: the Mapillary Geo-Localization (MGL) dataset \cite{c14} and the KITTI odometry dataset \cite{c41}. MGL provides diverse street-level images across various urban environments. We extract corresponding local OSM map crops using the noisy GPS priors provided by the MGL metadata. For KITTI, which focuses on autonomous driving scenarios, we utilize the Test2 split \cite{c14, c42} to ensure zero geographic overlap with the MGL training set. Local OSM map crops are sampled within a 20 meter radius of the attached GPS coordinates.

Semantic extraction utilizes DINOv2 with registers \cite{c43}. OSM elements are quantized into 7 area, 10 line, and 33 point categories. Map encoding is performed via a VGG16 based U-Net architecture. During the hierarchical estimation, the coarse stage employs a 1 meter spatial resolution and a 6$^\circ$ angular interval, while the fine stage refines the search space to 0.5 meters and 2$^\circ$. During training, spatial samples are drawn uniformly within a 32 meter radius of the ground truth. For MGL testing, GPS metadata is used only to define the queried 128$\times$128 meter local OSM map crop, and no pose prior is used within the map crop. For KITTI testing, we incorporate an initial pose prior within a $\pm 10^\circ$ window. Models are trained on three NVIDIA RTX 4090 GPUs, and all FPS results are measured on the same hardware.

\subsection{Comparison Methods}
We compare our framework with three categories of localization methods. For KITTI, we include satellite-based cross-view baselines, including DSM \cite{c46}, VIGOR \cite{c47}, and the refinement-based method in \cite{c42}. For MGL, we report retrieval-based and refinement-based monocular baselines \cite{c44, c27, c45}. Although newer OSM-based models such as \cite{c30} have been proposed, they are primarily optimized for multi-view autonomous-driving scenarios. Consequently, we prioritize comparison with OrienterNet, the open source state-of-the-art for monocular OSM localization.

\subsection{Performance on KITTI}
Evaluation on KITTI dataset (Table \ref{tab2}) demonstrates the robust generalization of our semantic representations to unseen geographic regions. We assess models trained exclusively on KITTI alongside those fine tuned from MGL pre training. Our framework consistently improves across all spatial and angular metrics when compared to both satellite- and OSM-based paradigms. Notably, when trained solely on KITTI, our method achieves a 3$^\circ$ orientation recall that exceeds the less stringent 5$^\circ$ recall of the prior leading approach.

\subsection{Performance on MGL}
Table \ref{tab1} details our performance on the MGL dataset. The proposed hierarchical strategy achieves substantial gains in both registration accuracy and inference efficiency. By dynamically pruning the search space, our full pipeline operates at 25.80 FPS on the same hardware, tripling the throughput of OrienterNet. Furthermore, the combination of DINO semantic alignment and dense local refinement enhances heading estimation, improving the rigorous 1$^\circ$ orientation recall by 38.48\% relative to existing benchmarks.

\section{Ablation Studies}

\begin{table}
\centering
\caption{Impact of DINO and coarse-to-fine (C2F) matching on localization performance.}
\label{tab3}
{
\begin{tabular}{lccccccc}
\toprule
\multirow{2}{*}{Method} & \multicolumn{3}{c}{Position R@Xm} & \multicolumn{3}{c}{Orientation R@X$^\circ$} & \multirow{2}{*}{FPS} \\ \cmidrule(lr{0pt}){2-4} \cmidrule(lr{0pt}){5-7}
             & 1m    & 3m    & 5m    & 1$^\circ$    & 3$^\circ$    & 5$^\circ$    &       \\ \midrule
Baseline    & 14.37 & 48.69 & 61.7  & 20.95 & 54.96 & 70.17 & 7.34  \\
+ DINO     & 17.06 & 53.17 & 65.31 & 24.54 & 61.28 & 76.35 & 6.81  \\
+ C2F      & \textbf{17.36} & \textbf{53.67} & \textbf{65.81} & \textbf{30.66} & \textbf{68.71} & \textbf{80.53} & \textbf{25.80}  \\
\bottomrule
\end{tabular}
}

\end{table}

To isolate the contributions of our core components, we establish a baseline utilizing a ResNet101 feature extractor coupled with global dense matching. As detailed in Table \ref{tab3}, integrating the DINO backbone significantly boosts pose recall, confirming its superiority in extracting reliable map-aligned semantics, albeit introducing a marginal computational overhead. The subsequent addition of the hierarchical matching strategy not only mitigates this overhead by increasing inference speed threefold, but also enables high-resolution angular sampling. This structural optimization yields clear improvements, particularly in strict orientation recall metrics.

\section{Conclusion}

We presented a hierarchical semantic alignment framework designed for monocular pose estimation within OpenStreetMap. By extracting globally aware object semantics via Vision Transformers, we explicitly bridged the severe modality gap existing between ground-level perspectives and overhead vector maps. Coupled with an uncertainty-aware search strategy, our approach achieves strong accuracy while satisfying the real-time computational constraints of autonomous systems. This research demonstrates the viability of utilizing lightweight vector maps for highly scalable and privacy preserving urban localization.


\addtolength{\textheight}{-2.2cm}

\bibliographystyle{IEEEtran}
\bibliography{ref}

\end{document}